\documentclass[12pt]{iopart}

\usepackage{xcolor}
\usepackage{mathrsfs}
\usepackage{amssymb}
\usepackage{pifont}
\usepackage{graphicx}

\usepackage{hyperref}
\usepackage{cite}

\usepackage{amsfonts}
\usepackage[T1]{fontenc}
\usepackage[utf8]{inputenc}

\begin{document}

\title[WADNet for the Characterization of Anomalous Diffusion]{WaveNet-Based Deep Neural Networks for the Characterization of Anomalous Diffusion (WADNet)}

\author{Dezhong Li, Qiujin Yao, Zihan Huang}

\address{School of Physics and Electronics, Hunan University, Changsha 410082, China}
\ead{huangzih@hnu.edu.cn}
\vspace{10pt}

\begin{abstract}
Anomalous diffusion, which shows a deviation of transport dynamics from the framework of standard Brownian motion, is involved in the evolution of various physical, chemical, biological, and economic systems. The study of such random processes is of fundamental importance in unveiling the physical properties of random walkers and complex systems. However, classical methods to characterize anomalous diffusion are often disqualified for individual short trajectories, leading to the launch of the Anomalous Diffusion (AnDi) Challenge. This challenge aims at objectively assessing and comparing new approaches for single trajectory characterization, with respect to three different aspects: the inference of the anomalous diffusion exponent; the classification of the diffusion model; and the segmentation of trajectories. In this article, to address the inference and classification tasks in the challenge, we develop a WaveNet-based deep neural network (WADNet) by combining a modified WaveNet encoder with long short-term memory networks, without any prior knowledge of anomalous diffusion. As the performance of our model has surpassed the current 1st places in the challenge leaderboard on both two tasks for all dimensions (6 subtasks), WADNet could be the part of state-of-the-art techniques to decode the AnDi database. Our method presents a benchmark for future research, and could accelerate the development of a versatile tool for the characterization of anomalous diffusion.
\end{abstract}

%
\vspace{2pc}
\noindent{\it Keywords}: anomalous diffusion, single trajectory characterization, anomalous diffusion challenge, machine learning, deep neural network, WaveNet
%
%
\maketitle
%
%

\section{Introduction}
A random walker undergoes anomalous diffusion when its mean squared displacement (MSD) deviates from a linear temporal evolution \cite{Metzler, Klafter}. Close attention is paid to such diffusion process in a variety of scientific fields, such as physics \cite{Mason,Aarao,Volpe,Xu}, chemistry \cite{Barkai}, biology \cite{Hofling,Wu,Gonzalez,Wang,Chen}, economics \cite{Plerou,Masoliver,Jiang}, and social science \cite{Castellano}. The raw information of anomalous diffusion dynamics can be recorded as trajectories of random walkers via techniques like single particle tracking \cite{Manzo,Shen,Qian,Saxton,Torreno-Pina}. Since these walkers directly probe their surroundings \cite{Wong,Banks,Chen2,Huang,Chen3,Lozano}, an in-depth characterization of their trajectory data is of significant importance in not only understanding the underlying mechanisms of anomalous diffusion, but also uncovering the intrinsic properties of both random walkers and complex environments.

Classical methods to analyze these trajectories are mostly based on mathematical statistics. A representative example is the calculation of MSD, which is quantified by the time-average or ensemble-average of squared displacements during a time interval \cite{Wang2,Kim}. These approaches can be valid to detect the diffusion properties when the trajectory is sufficiently long or a quantity of trajectories is available. However, as a consequence of the limitations of experimental techniques, one can usually obtain few, short, and noisy trajectories when recording a diffusion process \cite{Weron,Jeon,Akin}. These trajectories suffer the undesirable variations introduced by noise and a lack of significant statistics. Thus, mining information of anomalous diffusion from such a dataset are rather difficult and challenging for standard statistics-based approaches.

Therefore, alternative methods with improved performance on the characterization of anomalous diffusion are in great demand. This urgent issue leads to the launch of the Anomalous Diffusion (AnDi) Challenge (\url{http://www.andi-challenge.org}) which is spearheaded by a team of notable scientists \cite{Munoz-Gil}. The AnDi Challenge aims at assessing and comparing new approaches that go beyond classical methods for single trajectory characterization, regarding three tasks: the inference of the anomalous diffusion exponent; the classification of the diffusion model; and the segmentation of trajectories. Each task is split into 3 subtasks to deal with trajectories in different spatial dimensions (1D, 2D, and 3D). A challenge leaderboard is available by scoring the predictions of participating teams on a common test dataset. The rank of a team in the leaderboard can be an intuitive and objective criterion for the performance of corresponding method.

On the other hand, an unprecedented revolution of machine learning technologies has been witnessed in the past few years \cite{Silver,He,Hochreiter,Cho,Oord,Vaswani}. In particular, deep learning models that process sequential (or time-series) data are recently undergoing a rapid development \cite{Hochreiter,Cho,Oord,Vaswani}, from traditional recurrent neural networks (RNNs) \cite{Hochreiter, Cho} to the attention-based Transformer \cite{Vaswani}. These models can automatically learn the rules to extract useful information from sequences without any prior knowledge. Since trajectories of random walkers are actually sequences, these deep networks are highly expected to be qualified for the characterization of anomalous diffusion \cite{Bo,Munoz-Gil2,Munoz-Gil3,Gentili,Verdier,Argun,Manzo2}. In this article, as a response to the AnDi Challenge, we develop a WaveNet-based deep neural network (WADNet) by combining a modified WaveNet encoder \cite{Oord} with long short-term memory (LSTM) networks \cite{Hochreiter}, to address two tasks in the challenge: the inference of the anomalous diffusion exponent, and the classification of the diffusion model. Despite that no prior knowledge of anomalous diffusion is provided for WADNet, the performance of our model has surpassed the current 1st places in the challenge leaderboard on both two tasks for all dimensions (6 subtasks). Such a result suggests that WADNet could be the part of state-of-the-art techniques to decode the AnDi database.

\section{Methods}

\begin{figure}
\centering
\includegraphics[width=1.\textwidth]{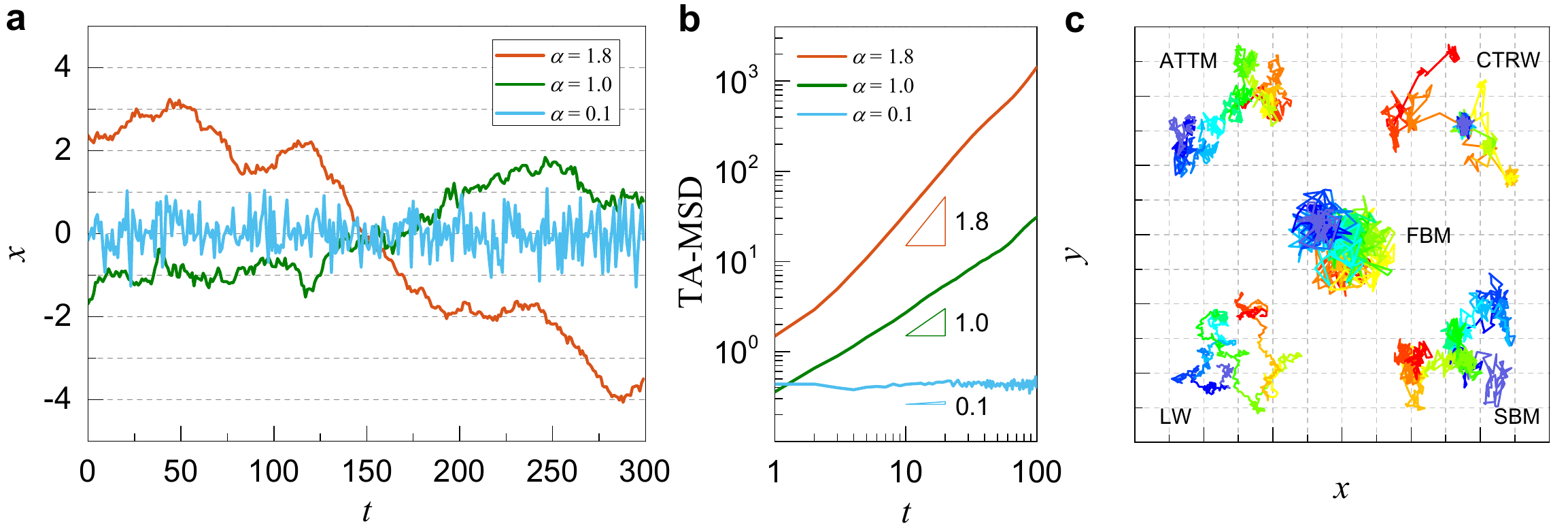}
\caption{{\bf Schematic illustrations of the inference and classification tasks in the AnDi Challenge.} \textbf{(a)} Representative 1D trajectories (part) of subdiffusion ($\alpha=0.1$), Brownian diffusion ($\alpha=1.0$), and superdiffusion ($\alpha=1.8$), respectively. \textbf{(b)} Time evolutions of TA-MSD corresponding to three trajectories in \textbf{(a)}. The exponent $\alpha$ can be determined by fitting the slope (shown as the triangle) of the log-log plot of TA-MSD versus time. \textbf{(c)} Representative trajectories of five diffusion models in the AnDi database: annealed transient time motion (ATTM), continuous-time random walk (CTRW), fractional Brownian motion (FBM), L\'{e}vy walk (LW), and scaled Brownian motion (SBM). The color on the trajectory denotes the time.}
\label{fig:fig1}
\end{figure}

\subsection{Tasks, evaluation metrics, and datasets}

In this work, we focus on two tasks in AnDi Challenge: the inference of the anomalous diffusion exponent (the inference task), and the classification of the diffusion model (the classification task). Brief introductions to these two tasks and corresponding evaluation metrics are given as follows:

\begin{itemize}
\item {\bf Inference of the anomalous diffusion exponent.}
One of the most common behaviors in anomalous diffusion dynamics is the nonlinear growth of MSD versus time, given by:
\begin{equation}
{\rm MSD}(t)\sim t^\alpha.
\end{equation}
Here, the anomalous diffusion exponent $\alpha$ is a real positive number. For standard Brownian diffusion we have $\alpha=1$, while for superdiffusion (subdiffusion) we have $\alpha>1$ $(\alpha<1)$. Representative 1D trajectories (part) of these three diffusion patterns are demonstrated in Figure 1a where $\alpha=0.1,1.0$, and $1.8$ respectively. To identify the exponent $\alpha$ of a single trajectory ${\bf x}(t)$, traditional statistics-based method is calculating the time-averaged mean squared displacement (TA-MSD), written as \cite{Munoz-Gil2}:
\begin{equation}
{\textrm {TA-MSD}}(t) = \frac{1}{t_{\rm max}-t}\int_0^{t_{\rm max}-t}
\|{\bf x}(\tau+t)-{\bf x}(\tau)\|_2^2{\rm d}\tau.
\end{equation}
For example, time evolutions of TA-MSD corresponding to three trajectories in Figure 1a are given in Figure 1b. As shown by the triangles, the exponent $\alpha$ can be determined by fitting the slope of the log-log plot of TA-MSD versus time. However, such an approach relies heavily on a sufficient long trajectory, and the result can usually be sensitive to the manual selection of fitting range. Hence, in this task, AnDi Challenge calls for improved methods to infer the anomalous diffusion exponent of a single short trajectory. The evaluation metric for this regression problem is the mean absolute error (MAE):
\begin{equation}
\mathrm{MAE}=\frac{1}{N} \sum_{i=1}^{N}\left|\alpha_{i, {\rm p}}-\alpha_{i, \mathrm{GT}}\right|,
\end{equation}
where $N$ is the number of trajectories in the test dataset, $\alpha_{i, {\rm p}}$ and $\alpha_{i, \mathrm{GT}}$ are the predicted and ground truth values of anomalous diffusion exponent of the $i$-th trajectory, respectively.

\item {\bf Classification of the diffusion model.}
It is well known that the properties of standard Brownian motion can be captured by the Wiener process. Analogously, the dynamics of anomalous diffusion can also be described by various theoretical stochastic process models. However, confident identification of these models from trajectory data is quite challenging for classical methods. Thus, to address this issue, the participants in this task are requested to develop valid approaches to classify the diffusion model of a single short trajectory. Five typical diffusion models are included in the challenge. Here, we give a brief introduction to these models and show the range of corresponding $\alpha$ in the AnDi database:
\begin{itemize}
\item
Annealed transient time motion (ATTM) \cite{Massignan}, a non-ergodic model that describes a Brownian motion with the diffusion coefficient randomly varying either in time or space ($0.05\leq\alpha\leq1$).
\item
Continuous-time random walk (CTRW) \cite{Scher}, a non-ergodic model where the waiting time of a random walker between two subsequent steps is irregular and randomly chosen ($0.05\leq\alpha\leq1$).
\item
Fractional Brownian motion (FBM) \cite{Mandelbrot}, an ergodic diffusion process that is driven by a fractional Gaussian noise. Such noise is normal distributed but has a power-law correlation ($0.05\leq\alpha<2$).
\item
L\'{e}vy walk (LW) \cite{Klafter2}, a non-ergodic model where the waiting time between subsequent steps is also irregular as CTRW but the step length is not Gaussian distributed ($1\leq\alpha\leq2$).
\item
Scaled Brownian motion (SBM) \cite{Lim}, a non-ergodic model that describes a Brownian motion with a deterministically time-dependent diffusion coefficient ($0.05\leq\alpha\leq2$).
\end{itemize}
Detailed descriptions regarding the mechanisms of these theoretical models can be found in the instruction of AnDi Challenge \cite{Munoz-Gil}. Representative trajectories of these five models are shown in Figure 1c, where color on the trajectory denotes the time. The evaluation metric in this classification task is the "micro" version of F1-score, expressed as:
\begin{equation}
{\rm F1} = 2 \cdot \frac{\rm { precision } \cdot \rm { recall }}{\rm { precision }+\rm { recall }}.
\end{equation}
Here, $ {\rm precision}=\frac{\mathrm{TP}}{\mathrm{TP}+\mathrm{FP}}$ and ${\rm recall }=\frac{\mathrm{TP}}{\mathrm{TP}+\mathrm{FN}}$, where TP, FP and TN represent the {\it total} number of true positives, false positives and false negatives in the classification result, respectively.
\end{itemize}

\begin{table}
\caption{\label{jlab1}List of specific lengths and corresponding trajectory numbers in training set.}
\centering
\footnotesize
\begin{tabular*}{\textwidth}{@{}c*{6}{@{\extracolsep{10pt plus10pt}}c}}
\br
Length & Number & Length & Number & Length & Number \\
\mr
10& 5000000&	105&	2000000&	400&	1000000\\
15&	5000000&	115&	2000000&	425&	1000000\\
20&	5000000&	120&	2000000&	450&	1000000\\
25&	5000000&	125&	2000000&	475&	1000000\\
30&	5000000&	150&	2000000&	500&	1000000\\
40&	5000000&	175&	2000000&	550&	1000000\\
45&	5000000&	200&	2000000&	600&	1000000\\
50&	5000000&	225&	2000000&	650&	1000000\\
55&	5000000&	250&	2000000&	700&	1000000\\
60&	5000000&	250&	2000000&	750&	1000000\\
70&	5000000&	275&	1500000&    800&	1000000\\
80&	5000000&	300&	1500000&    850&	1000000\\
90&	5000000&	325&	1500000&    900&	1000000\\
100&5000000&	350&	1500000&    950&	1000000\\
105&2000000&    375&    1500000& & \\
\br
\end{tabular*}
\end{table}
\normalsize

The datasets in the AnDi Challenge are composed of simulated trajectories that are generated by an open-source Python package {\it andi-dataset} (\url{https://github.com/AnDiChallenge/ANDI_datasets}) \cite{Gorka}. The working principle of {\it andi-dataset} is briefly summarized here. Trajectory generation is based on the theoretical framework of five diffusion models in the database, where time steps are uniformly sampled with a unitary time interval. The ground truth value of anomalous diffusion exponent $\alpha_{\rm GT}$ ranges from 0.05 to 2.0 with a step 0.05. Trajectory length varies from 10 to 999 points, where one point denotes the spatial coordinate of the random walker at one time step. In particular, to mimic real experimental data, simulated trajectories are corrupted with zero-mean Gaussian noise. The signal-to-noise ratio (SNR) of a trajectory can be regulated through changing the standard deviation $\sigma_n$ of noise distribution (SNR $= 1/\sigma_n$). In the {\it andi-dataset}, $\sigma_n$ is set as 1.0, 0.5, or 0.1, resulting in SNR = 1, 2, or 10.

On the basis of {\it andi-dataset}, a test dataset consisting of 10000 trajectories is generated by organizers as the common dataset for leaderboard scoring. The training dataset we utilized in this work is also generated by the package {\it andi-dataset}. For both the inference and classification tasks, we choose 43 specific lengths and separately generate the training subsets at these lengths. The values of these specific lengths and corresponding trajectory numbers are listed in Table 1. Before training or inference, trajectory data is normalized to ensure its position's average and standard deviation in each dimension are 0 and 1 respectively.

\subsection{Workflow of WADNet}

The overview of the entire workflow of WADNet is demonstrated in Figure 2, including three main parts: {\it Input}, {\it WaveNet Encoder}, and {\it RNN} \& {\it MLP}. As shown in the {\it Input} part, input data can be obtained through a simple preprocessing of the raw trajectory, which is valid for trajectories in arbitrary dimensions: i. we separate the trajectory in each dimension into 1D time-series data; ii. we stack the 1D time-series data of all dimensions to compose a multi-channel tensor as the input of WADNet. For instance, the transformation of a 3D trajectory ${\bf x}(t) = [x_1(t),x_2(t),x_3(t)]^{\rm T}$ with a length $T$ to an input tensor $\bar{\bf x}$ can be expressed as:

\begin{equation}
\fl \left\{\begin{array}{c}
\left[\begin{array}{c}
x_{1}(1) \\
x_{2}(1) \\
x_{3}(1)
\end{array}\right],  \left[\begin{array}{c}
x_{1}(2) \\
x_{2}(2) \\
x_{3}(2)
\end{array}\right],   \cdots,  \left[\begin{array}{c}
x_{1}(T) \\
x_{2}(T) \\
x_{3}(T)
\end{array}\right]
\end{array}\right\}\rightarrow\left\{\begin{array}{l}
{\left[x_{1}(1), x_{1}(2),  \cdots, x_{1}(T)\right]} \\
{\left[x_{2}(1), x_{2}(2),  \cdots, x_{2}(T)\right]} \\
{\left[x_{3}(1), x_{3}(2),  \cdots, x_{3}(T)\right]}
\end{array}\right\}
\end{equation}

After the preprocessing, the input tensor is going to pass through a modified WaveNet encoder. WaveNet is a deep neural network designed by Google Deepmind for the purpose of raw audio generation \cite{Oord}. The encoder of WaveNet based on dilated causal convolutions, gated activation units, and skip connections has been proved to be a powerful tool for extracting information from sequential data \cite{Kleijn, Chorowski}. In the {\it WaveNet Encoder} part of Figure 2, we show the detailed architecture of the modified WaveNet encoder utilized in this work. The input tensor $\bar{\bf x}$ is firstly filtered by a convolutional layer with filter size $1\times 3$ and filter number 32. Corresponding output $\tilde{\bf x}_{1}$ is subsequently processed by the stacked dilated convolutions with a dilation depth $d = 16$. Here, a dilated convolution with a dilation $m$ is a convolution operation where the filter skips the input values with a step $m-1$. Such a structure allows the filter to deal with a larger area than its size. For example, a $3\times 3$ filter with a dilation 4 can be directly applied on a $9\times9$ area. Thus, by simply stacking the dilated convolutions, networks are enabled to have very large receptive fields with just a few layers. In this work, the dilation is doubled for every layer, which is the same as the original WaveNet \cite{Oord}.

\begin{figure}
\centering
\includegraphics[width=1\textwidth]{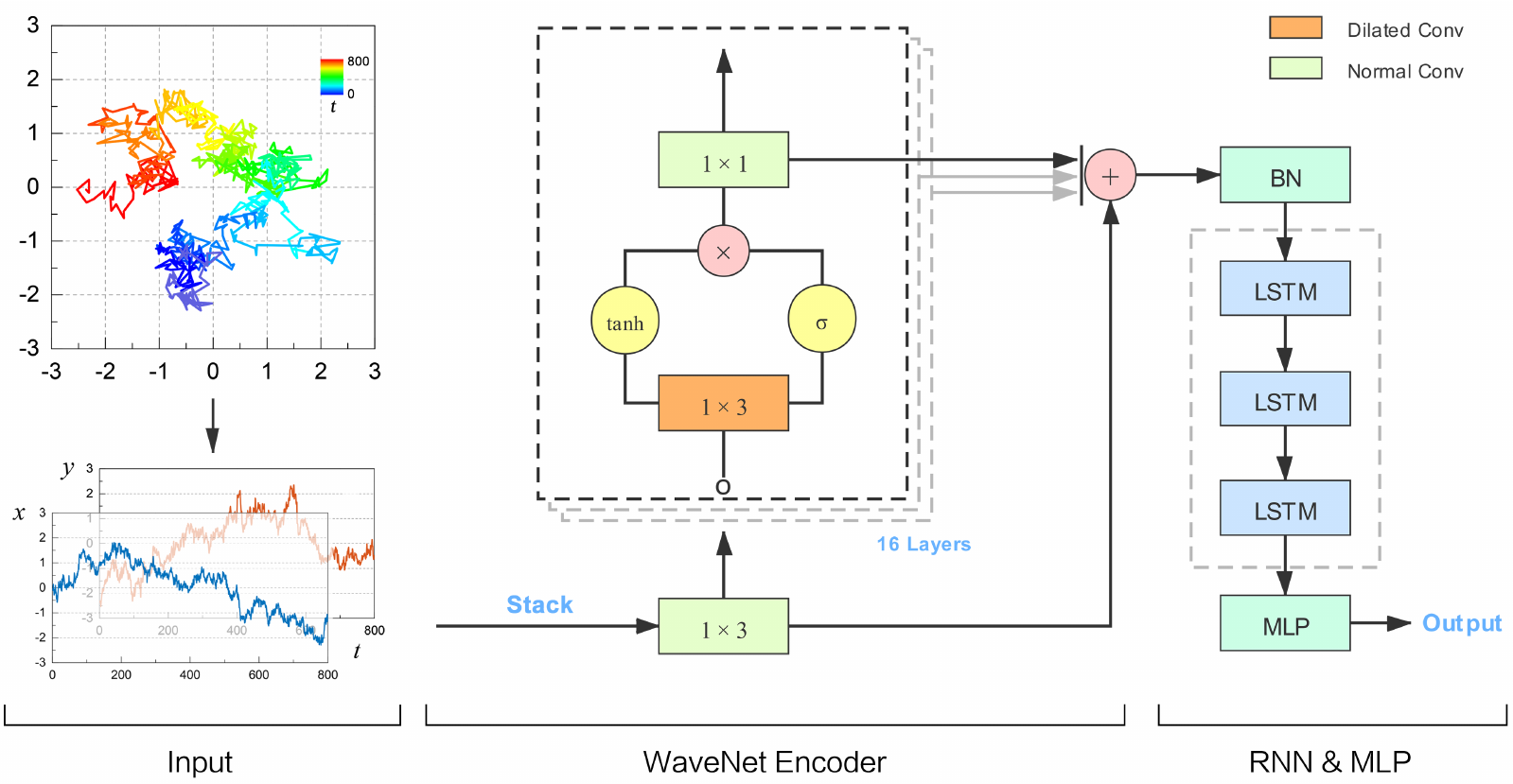}
\caption{\textbf{Overview of the workflow of WADNet.}  The workflow of WADNet consists of three main parts: {\it Input}, {\it WaveNet Encoder}, and {\it RNN} \& {\it MLP}. The {\it Input} part describes the preprocessing method of raw trajectories. The {\it WaveNet Encoder} part shows the detailed architecture of the modified WaveNet encoder. The output of this encoder is utilized as the input of the following {\it RNN} \& {\it MLP} part, in which the inference and classification tasks are accomplished.}
\label{fig:fig2}
\end{figure}

Next, to apply the gated activation unit in each layer, dilated convolution in the WaveNet encoder is composed of two parallel convolution operators with same structures, named as "filter" and "gate". The outputs of "filter" and "gate" are activated by the hyperbolic tangent and sigmoid functions respectively, leading an output ${\bf z}_k$ of the gated activation unit:
\begin{equation}
{\bf z}_k = \tanh\left(W_{f,k}\ast\tilde{\bf x}_k\right)\otimes\sigma\left(W_{g,k}\ast\tilde{\bf x}_k\right).
\end{equation}
Here, $k$ is the layer index, $\tilde{\bf x}_k$ is the input tensor of the $k$-th layer, $W$ represents the dilated convolution filter with filter size $1\times 3$ and filter number 32, $f$ and $g$ denote the "filter" and "gate", $\ast$ denotes the convolution operator, $\tanh(\cdot)$ is the hyperbolic tangent activation function, $\sigma(\cdot)$ is the sigmoid activation function, $\otimes$ denotes the element-wise multiplication operator. After that, a $1\times1$ convolution filter $W_1$ with filter number 32 is applied on ${\bf z}_k$ to obtain the input of next layer, written as:
\begin{equation}
\tilde{\bf x}_{k+1} = W_1\ast {\bf z}_k, \quad k=1,2,3,\cdots,d.
\end{equation}
In particular, this operation is different from that in the original WaveNet encoder, where residual \cite{He} is used in the layer, i.e., $\tilde{\bf x}_{k+1} = W_1\ast {\bf z}_{k} + \tilde{\bf x}_{k}$. The output tensor ${\bf u}$ of this modified encoder is obtained through skip connections:
\begin{equation}
{\bf u} = {\rm BN}\left(\sum\nolimits_{k = 1}^{d + 1}{\tilde{\bf x}_{k}}\right),
\end{equation}
where ${\rm BN}(\cdot)$ represents the 1D batch normalization.

At last, ${\bf u}$ is used as the input of the following RNN module. As show in the {\it RNN} \& {\it MLP} part of Figure 2, this module is made up of 3 stacked LSTM units, whose dimension of hidden layers is 64. The output ${\bf v}$ of RNN module is the feature vector of original trajectory learned by WADNet. To accomplish the challenge task, feature ${\bf v}$ is processed by a multilayer perceptron (MLP) network. For the inference task, the output dimension of MLP is 1 and no activation function is utilized; for the classification task, the output dimension of MLP is 5 and the activation function is the softmax function.

\subsection{Training procedure and inference strategy}

Models for each specific length are trained separately. 80\% of training data is used for training, while the other is used for validation. The model is trained using back-propagation with a batch size 512, where loss function is the mean squared error (MSE) and cross entropy (CE) for the inference and classification tasks, respectively. The optimizer is Adam with an initial learning rate $l=0.001$. During the training process, the learning rate is changed as
$l_{\rm new} = l_{\rm old}/5 $
if the valid loss doesn't decrease after 2 epochs. When the number of such learning-rate changes exceeds 2, the training process is early stopped to save training time and avoid overfitting.

To demonstrate the convergence of our method, we take the 1D trajectory at length 300 as the example, and show corresponding evolutions of train loss, valid loss, and valid metric in Figures 3a (the inference task) and 3b (the classification task). Since train and valid losses can both be effectively reduced during training, an excellent convergence of WADNet can be identified. In particular, we determine the best model weight for a specific length by the best valid metric, as depicted by the arrows in Figures 3a and 3b. Moreover, to examine the model performance in the training process, we show the best valid metrics of all specific lengths in all dimensions in Figures 3c (the inference task) and 3d (the classification task). For both two tasks, the performance of WADNet is better for higher spatial dimensions, and is gradually boosted as the trajectory length increases. This result arises from the fact that a longer trajectory with a higher dimension can provide more data points and hidden information.

\begin{figure}
\centering
\includegraphics[width=1\textwidth]{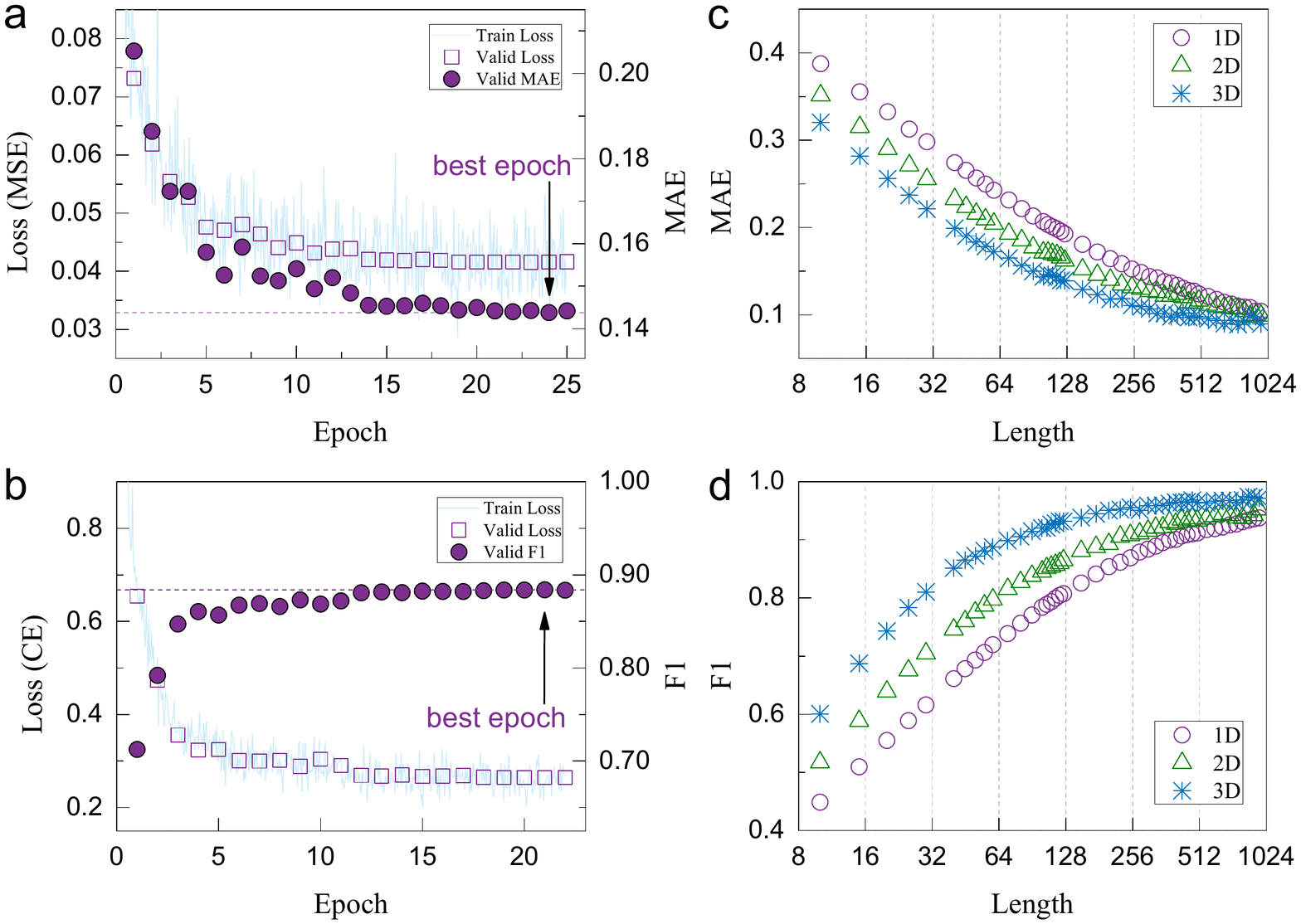}
\caption{{\bf Model performance in the training process.} {\bf (a-b)} Evolutions of train loss, valid loss, and valid metric for the 1D trajectory at length 300, regarding the inference task {\bf (a)} and the classification task {\bf (b)}. The arrow denotes the best model weight for a specific length, which is determined by the best valid metric. {\bf (c-d)} The best valid metrics of all specific lengths in all dimensions for the inference task {\bf (c)} and the classification task {\bf (d)}.}
\label{fig:fig3}
\end{figure}

Next, we pay attention to the inference of test dataset. Note that the length of a trajectory in the test dataset varies from 10 to 999 points. Since we only have model weights for 43 specific lengths, our model cannot be directly applied for inference. To address this issue, the inference of test dataset is guided by the following strategy:
\begin{itemize}
\item If the original length of a trajectory belongs to these 43 specific lengths, the data of this trajectory will be directly used for inference.
\item Otherwise, a new length of this trajectory will be set as the closest smaller specific length. For instance, the new length of a trajectory with an original length 49 should be 45. The trajectory data is subsequently transformed into 2 sequences. For clarity, we set the trajectory as ${\bf x} = [x_1, x_2, \cdots, x_T]$, where $T$ is the original length. We denote $T_n$ as the new length. Note that $T_n < T$, two sequences ${\bf x}_1 = [x_1, x_2, \cdots, x_{T_n}]$ and ${\bf x}_2 = [x_{T-T_n+1}, x_{T-T_n+2}, \cdots, x_T]$ with length $T_n$ can be obtained. Such two sequences are both used for inference, where corresponding feature vectors learned by WADNet are ${\bf v}_1$ and ${\bf v}_2$ respectively. After that, we choose the mean vector of these two vectors as the approximation of the feature ${\bf v}$ of the original trajectory, i.e., ${\bf v}\approx ({\bf v}_1+{\bf v}_2)/2$. The final prediction is made based on ${\bf v}$ by the MLP.
\end{itemize}
The codes of WADNet for training and inference are freely available online \cite{Li}, where the neural networks are implemented by PyTorch 1.6.0.

In addition, to further improve the performance of WADNet, $K$-fold cross validation technique \cite{Stone} is utilized for both two tasks. In detail, the training dataset is randomly split into $K$ folds where $K=5$ in this work. When training a model, one fold is utilized as the validation data while other folds are used as training data. Such process is then repeated $K$ times, with each of the folds used exactly once as the validation data. Thus, the prediction of a sample can be obtained by averaging the results of $K$ single-fold models. On the other hand, for the inference of the anomalous diffusion exponent, predictions of test dataset are all rectified by multiplying a constant $w$ and are subsequently clipped to ensure $\alpha \in [0.05, 2.0]$. The constant $w$ is determined by analyzing an external validation dataset containing 100000 trajectories, and is identified to be $1.0110, 1.0129$, and $1.0083$ for 1D, 2D, and 3D trajectories, respectively.

\section{Results and Discussions}

\begin{figure}
\centering
\includegraphics[width=1\textwidth]{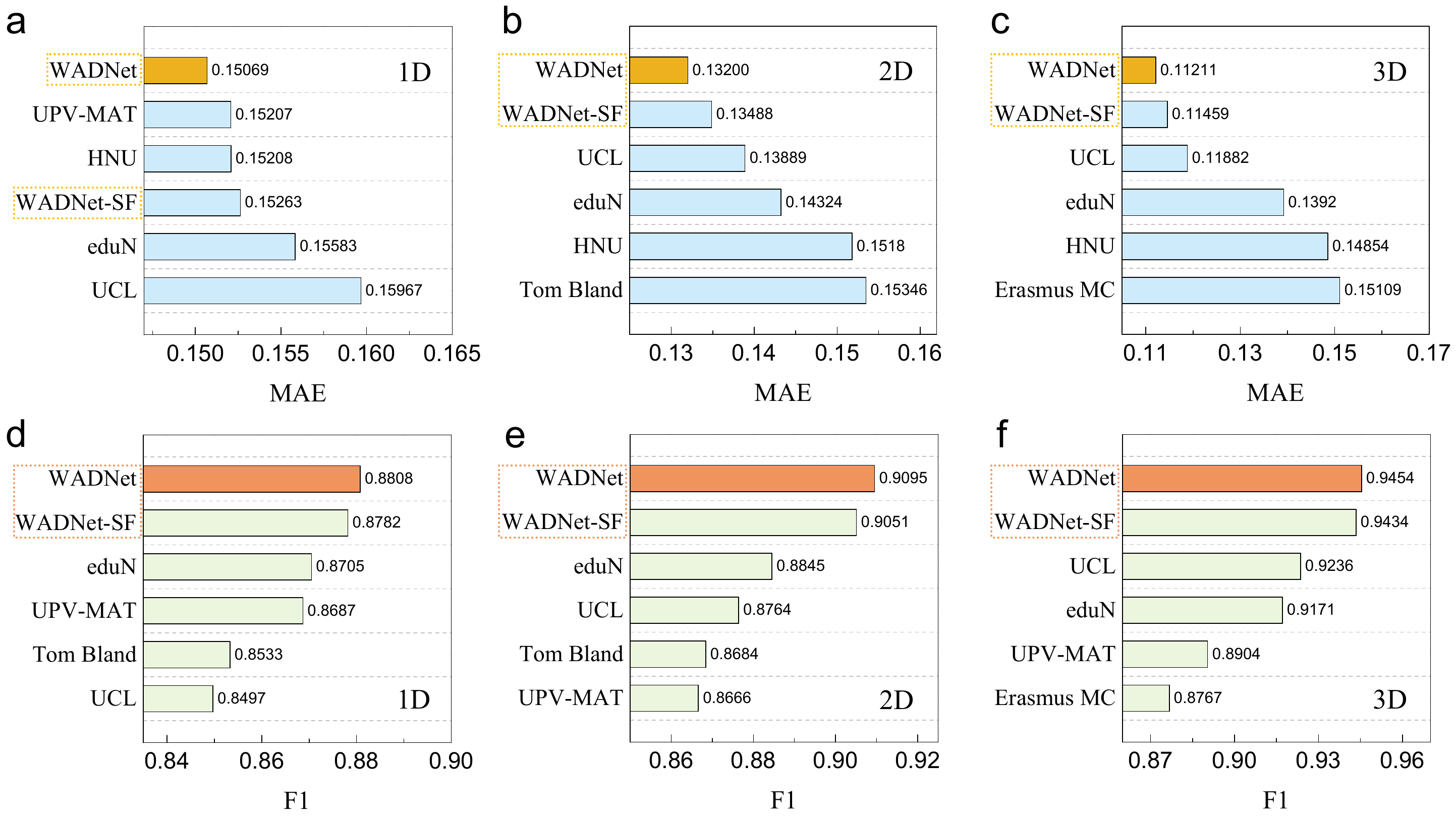}
\caption{\textbf{Performance of WADNet in the challenge leaderboard.} Results of comparisons between WADNet and the top 4 ranks in the challenge leaderboard. Label "WADNet" denotes the score of our method improved by the $K$-fold cross validation, and label "WADNet-SF" denotes the score of a single-fold WADNet model. The best score in each subtask is marked in different color for clarity. \textbf{(a-c)} MAE scores in the inference task for 1D {\bf (a)}, 2D {\bf (b)}, and 3D {\bf (c)} trajectories. \textbf{(d-f)} F1-scores in the classification task for 1D {\bf (d)}, 2D {\bf (e)}, and 3D {\bf (f)} trajectories.}
\label{fig:fig4}
\end{figure}

\subsection{Performance of WADNet in the challenge leaderboard}

To examine objectively the performance of WADNet in the AnDi Challenge, we compare our scores of test dataset with the top 4 ranks in the challenge leaderboard. The results of comparisons for both the inference and classification tasks in all dimensions (6 subtasks) are summarized in Figure 4, where the best scores are marked in different colors for clarity. It can be found that the scores of our method have surpassed the current 1st places in the leaderboard in all 6 subtasks. In particular, without the improvement of $K$-fold cross validation, the scores of a single-fold model (WADNet-SF) have also surpassed the 1st places in 5 subtasks. This result illustrates that the ability of WADNet in the characterization of anomalous diffusion mainly arises from the model itself. Therefore, WADNet could be the part of state-of-the-art techniques for the AnDi database. Note that there are only 43 specific lengths in the training dataset. That is, by a simple expansion of training data, the performance of WADNet can be further improved. Analysis in the following sections is based on the best performance of WADNet that is improved by the $K$-fold cross validation.

\subsection{Analysis of the inference of the anomalous diffusion exponent}

\begin{figure}
\centering
\includegraphics[width=1\textwidth]{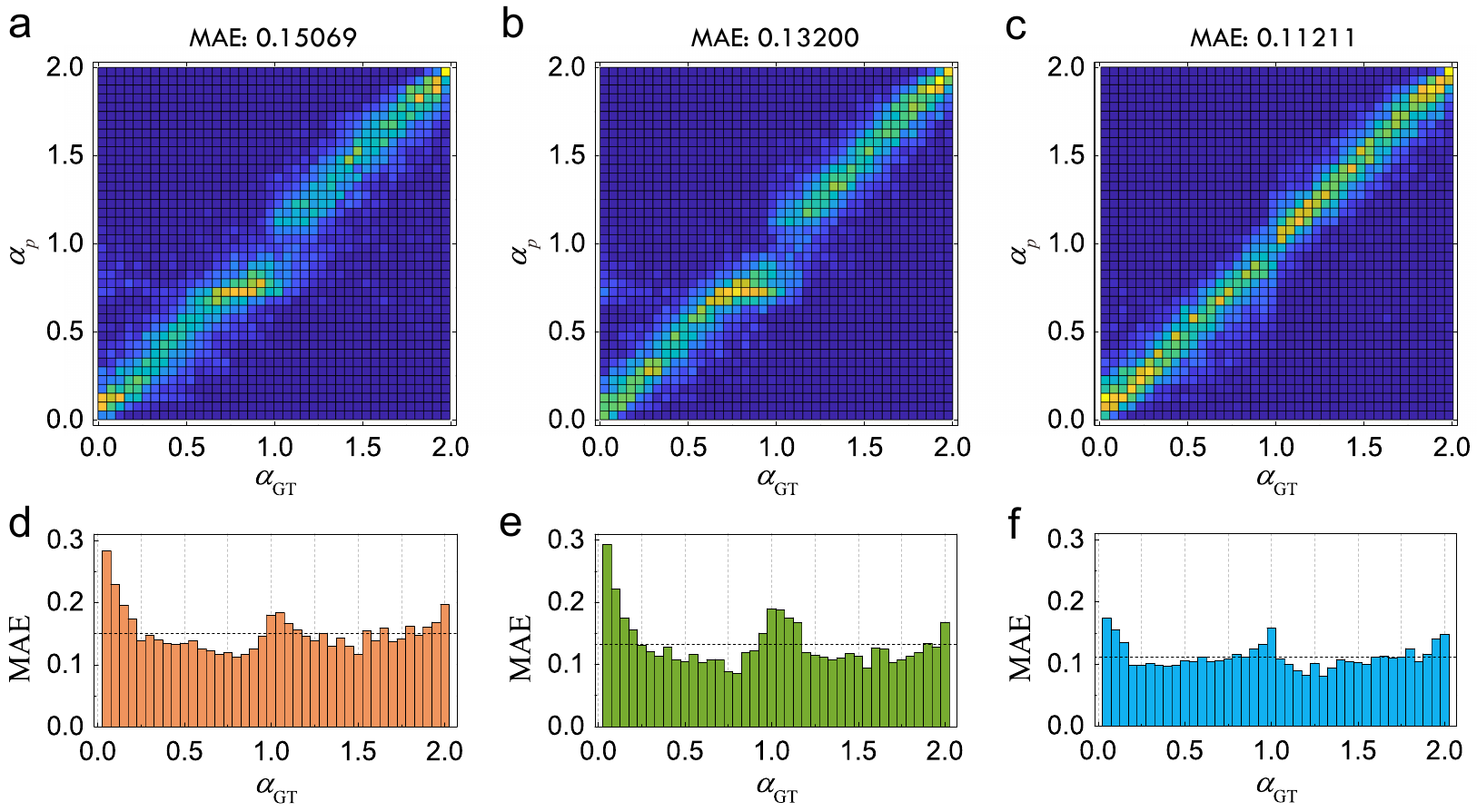}
\caption{\textbf{Differences between the predicted values and ground truth values of the anomalous diffusion exponent.} \textbf{(a-c)} 2D histograms of $\alpha_{\rm p}$ versus $\alpha_{\rm GT}$ for 1D {\bf (a)}, 2D {\bf (b)}, and 3D {\bf (c)} trajectories. \textbf{(d-f)} Plots of MAE versus $\alpha_{\rm GT}$ for 1D {\bf (d)}, 2D {\bf (e)}, and 3D {\bf (f)} trajectories. The dashed lines denote the mean values of MAE.}
\label{fig:fig5}
\end{figure}

In this section, we provide a detailed analysis of the performance of WADNet on the inference task. Firstly, as the spatial dimension of trajectory increases, MAE value of test dataset is reduced from 0.15069 (1D) to 0.13200 (2D) to 0.11211 (3D). This result can be attributed to the extra information given by a higher dimension. Moreover, to intuitively demonstrate the differences between the predicted values $\alpha_{\rm p}$ and ground truth values $\alpha_{\rm GT}$, we plot the 2D histograms of $\alpha_{\rm p}$ versus $\alpha_{\rm GT}$ for all dimensions in Figures 5a-c. It can be found that 1D and 2D trajectories with $\alpha_{\rm GT} \approx 1.0$ are more likely to be predicted as subdiffusion ($\alpha_{\rm p} < 1$), while such a deviation is slightly eliminated in 3D space. This bias can also be captured by the plots of MAE versus $\alpha_{\rm GT}$ (Figures 5d-f), where MAE values at $\alpha_{\rm GT} \approx 1.0$ are relatively higher than that of neighborhoods. Meanwhile, significant errors can be identified at $\alpha_{\rm GT}\rightarrow 0$. The reason is that a random walker in this case is nearly motionless and corresponding trajectory is more like a record of localization noise.

Next, we explore the dependence of model performance on different diffusion models, and show the results in Figure 6a. As expected, FBM that is ergodic has the lowest MAE values in all dimensions. Interestingly, for those non-ergodic models, WADNet shows a better performance on CTRW and LW than on ATTM and SBM, indicating that trajectories of CTRW and LW could provide more hidden information than the other two. Additionally, MAE values of CTRW and LW are nearly independent of the trajectory dimension, while those of ATTM, FBM, and SBM decrease as the dimension increases. This result suggests that the quality of features of CTRW and LW learned by WADNet is less correlated with the spatial dimension.

\begin{figure}
\centering
\includegraphics[width=1.\textwidth]{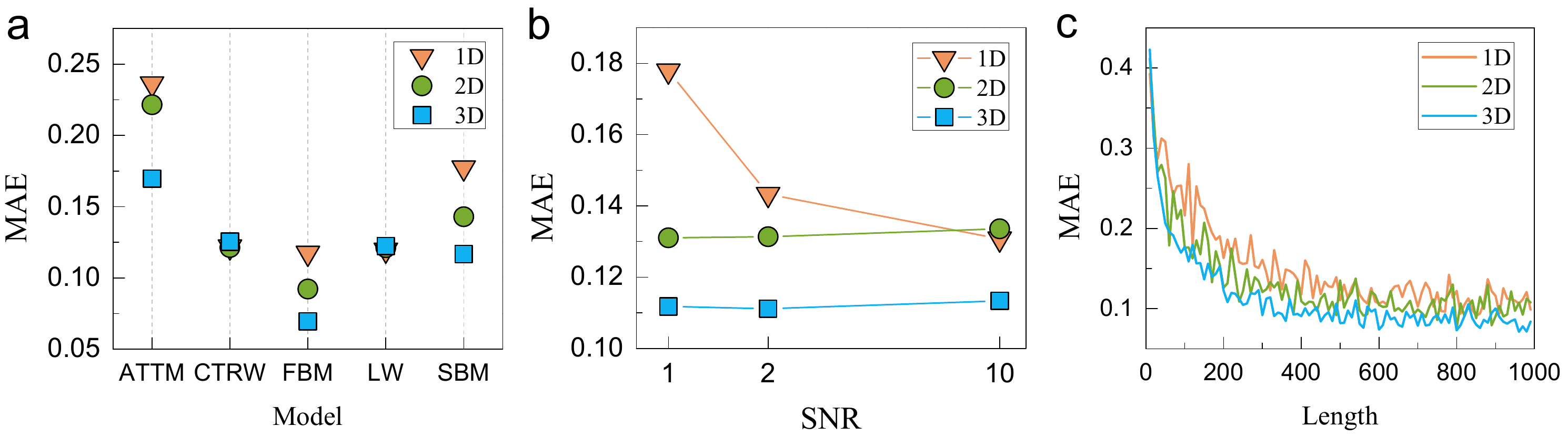}
\caption{\textbf{Influence of the diffusion model, SNR, and trajectory length on the inference task.} {\textbf (a)} MAE of 5 diffusion models. {\textbf (b)} Plots of MAE versus SNR. {\textbf (c)} Dependence of MAE on the trajectory length.}
\label{fig:fig6}
\end{figure}

On the other hand, since experimental trajectories can usually be noisy, it is of crucial importance to examine the performance of WADNet on noisy trajectories. For that purpose, we investigate the effects of SNR on the performance of our model. As shown in Figure 6b, MAE value can be slightly reduced by the improvement of SNR for the 1D trajectory. However, a surprising result is that MAE values are nearly the same for all SNR in 2D and 3D spaces. Such a result implies that WADNet has the ability to remove the undesirable effects introduced by localization noise for 2D and 3D trajectories, even for trajectories where the standard deviation of noise has the same amplitude as that of displacement (i.e., SNR = 1). Therefore, our method is of high potential to be utilized for the analysis of noisy experimental trajectory data. At last, the influence of trajectory length is demonstrated in Figure 6c. As expected, MAE rapidly decreases with the increase of trajectory length, and reaches a relatively steady value near the length 600.

\begin{figure}
\centering
\includegraphics[width=1.\textwidth]{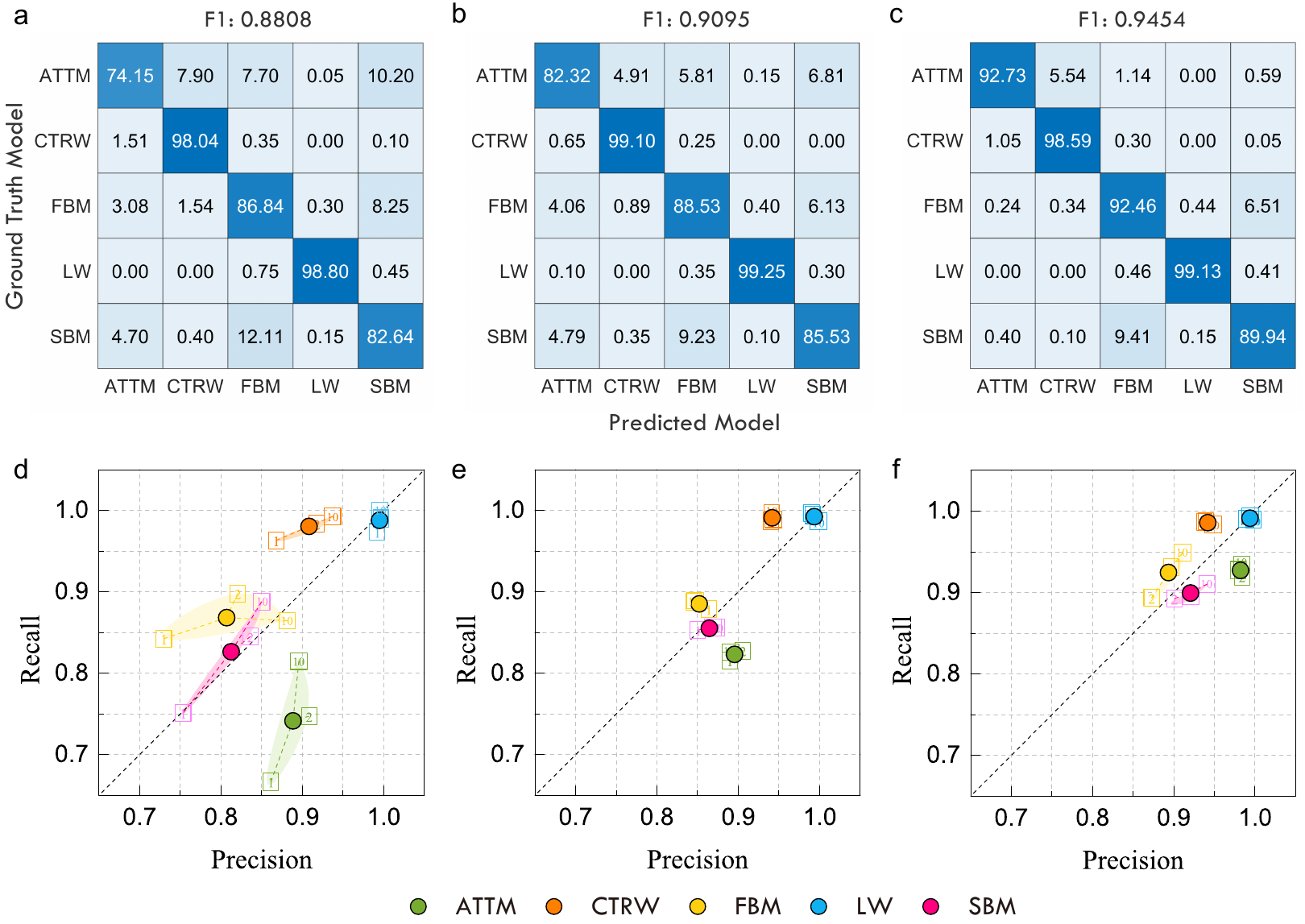}
\caption{\textbf{Detailed analysis of the performance of WADNet on the classification task.} {\textbf (a-c)} Confusion matrixes of classification results for 1D {\bf (a)}, 2D {\bf (b)} and 3D {\bf (c)} trajectories. The element of confusion matrix is the percentage, and the diagonal element of matrix is the classification accuracy for each model. {\textbf (d-f)} Plots of precision versus recall for 1D {\bf (d)}, 2D {\bf (e)} and 3D {\bf (f)} trajectories. The dots denote the overall scores, and the squares denote the results for different SNR. The colored region is a guide of square locations to the eye, and its area reports the stability of WADNet on noisy trajectories for each model.}
\label{fig:fig7}
\end{figure}

\subsection{Analysis of the classification of the diffusion model}

In this section, we focus attention on the performance of WADNet on the classification task. At first, F1-score of test dataset is boosted from 0.8808 (1D) to 0.9095 (2D) to 0.9454 (3D), as a consequence of the information gain given by a higher spatial dimension. To examine the classification performance of WADNet in detail, we show the confusion matrixes for 1D, 2D and 3D trajectories in Figures 7a-c respectively. For clarity, we use the percentage instead of sample number as the element of confusion matrix. Thus, the diagonal element of matrix is actually the classification accuracy of WADNet for each model. As characterized by the accuracy, the performance on CTRW and LW is relatively better (accuracy > 98\% in all dimensions) and almost independent on the spatial dimension, while the performance on ATTM, FBM, and SBM is gradually improved as the dimension increases.

From another perspective, the classification ability of WADNet is further analyzed by calculating the precision and recall scores of each diffusion model. The plots of precision versus recall for all dimensions are given in Figures 7d-f as dots, where the diffusion models are distinguished by different colors. A general phenomenon in all dimensions is that precision and recall are approximately equivalent for LW and SBM, but show non-negligible differences for the other 3 models. In more detail, the recall is smaller than precision for ATTM, and undergoes a reverse for FBM and CTRW. That is, regarding the error of classifying a trajectory, ATTM is more likely to be mistaken, while FBM and CTRW are more preferred to be the targets.

Next, similar with the inference task, the performance of WADNet on noisy trajectories is also examined by studying the effects of SNR. Corresponding results in all dimensions are plotted in Figures 7d-f as squares marked by SNR values. The colored region is a guide of square locations to the eye, and its area reports the stability of our method on noisy trajectories for each model. As characterized by the small area of colored regions, insensitivity of F1-score to SNR can be identified for LW and CTRW in all dimensions. In contrast, F1-scores of ATTM, FBM, and SBM in 1D space can be slightly improved by the increase of SNR. However, for 2D and 3D trajectories, F1-scores of these 3 models are nearly independent of SNR, as the same with LW and CTRW. Thus, negative influences of localization noise on the classification performance could be vastly eliminated by WADNet.

\begin{figure}
\centering
\includegraphics[width=0.93\textwidth]{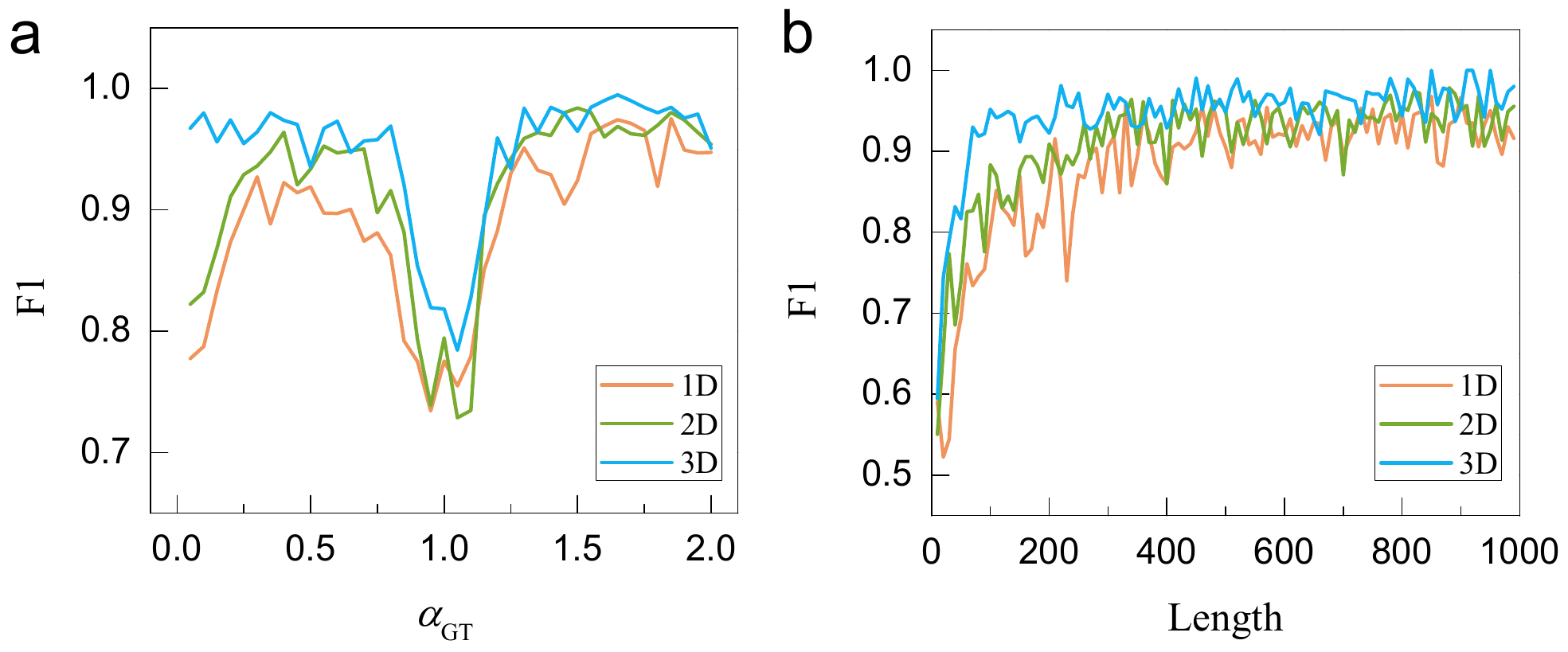}
\caption{\textbf{Influence of anomalous diffusion exponent and trajectory length on the classification task.} {\bf (a)} Plots of F1-score as a function of $\alpha_{\rm GT}$. {\bf (b)} Plots of F1-score versus trajectory length.}
\label{fig:fig8}
\end{figure}

Moreover, to investigate the influence of anomalous diffusion exponent on the classification result, we show the F1-score as a function of $\alpha_{\rm GT}$ in Figure 8a. Obviously, the worst performance of WADNet can be identified for trajectories with $\alpha_{\rm GT}\approx 1.0$ in all dimensions. The possible reason is that the dynamics of these trajectories is close to Brownian motion, which leads to similar statistical properties regardless of the specific diffusion model. In addition, the classification performance is also poor at $\alpha_{\rm GT}\rightarrow 0$ in 1D and 2D spaces. This result can be attributed to the fact that the transport of a random walker with an extremely small $\alpha$ is nearly immobile and dominated by the random noise. This phenomenon vanishes for 3D trajectories, probably owing to the information gain. In addition, the dependence of F1-score on trajectory length is given in Figure 8b. As was anticipated, F1-score undergoes a rapid growth as the trajectory length increases, and reaches a relatively steady value near the length 600.

At last, to gain a more in-depth insight into the classification ability of our method, we present a visualization of feature vector {\bf v} extracted from the penultimate layer of WADNet. Figure 9 shows the uniform manifold approximation and projection for dimension reduction (UMAP) \cite{Mclnnes} projections of feature vectors of 1D trajectories with length 40, 200, 400, and 900, where each length has 10000 trajectories. Projections of features are marked by different colors according to corresponding diffusion models. It can be found that features are mixed for short trajectories, but can be clearly distinguished for long trajectories. In more detail, as the trajectory length increases, the features of LW can be firstly separated (length 200), and two main isolated clusters can be identified next (length 400 and 900). One cluster is primarily composed of FBM and SBM, and the other is a combination of CTRW and ATTM. Therefore, from the perspective of feature similarity, the underlying dynamics of these diffusion models is not completely uncorrelated. The overlap between FBM and SBM suggests that the diffusion driven by a fractional Gaussian noise can sometimes be similar with the Brownian motion with a time-dependent diffusivity. Meanwhile, the effects of an irregular waiting time might be approximated by a random variation of diffusivity at times, as indicated by the adjacency between CTRW and ATTM. For future research, these features learned by WADNet have the potential to be used as quantitative representations of trajectories with multiple and mixed diffusion dynamics.

\begin{figure}
\centering
\includegraphics[width=1.\textwidth]{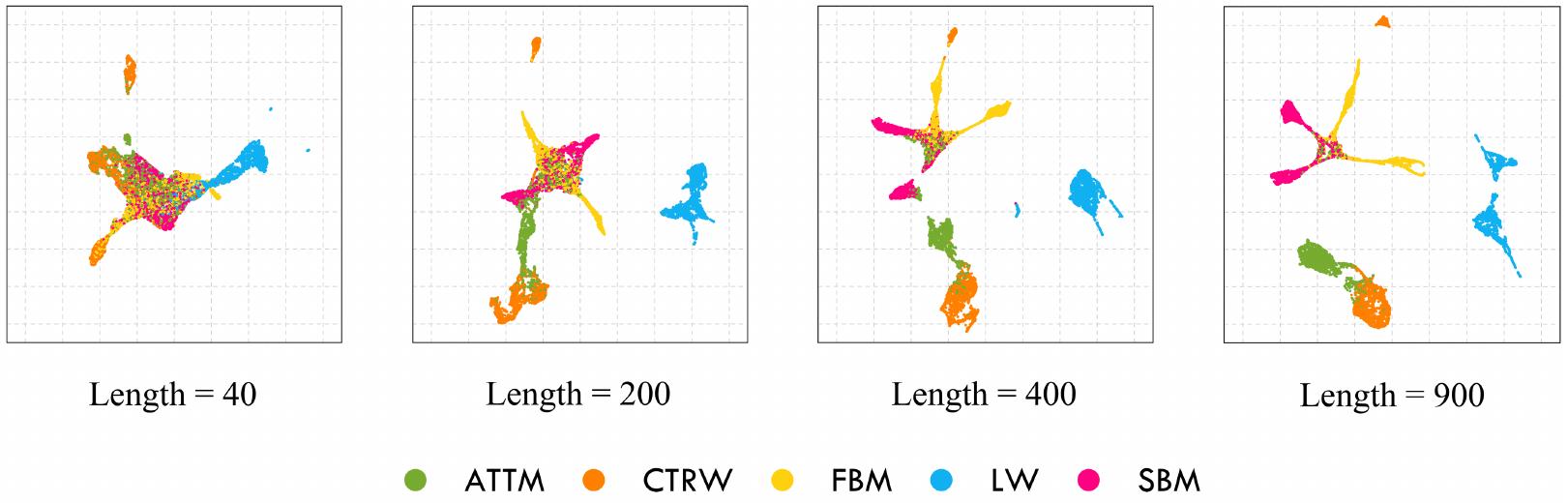}
\caption{\textbf{Visualization of learned features.} Uniform manifold approximation and projection for dimension reduction (UMAP) projections of feature vectors of 1D trajectories with length 40, 200, 400, and 900, where each length has 10000 trajectories. Feature vectors are extracted from the penultimate layer of WADNet.}
\label{fig:fig8}
\end{figure}

\section{Conclusions}
In conclusion, by combining a modified WaveNet encoder with LSTM networks, we have developed a WaveNet-based deep neural network (WADNet) for the characterization of anomalous diffusion in response to the AnDi Challenge. Without any prior knowledge of anomalous diffusion, the performance of WADNet has surpassed the current 1st places in the challenge leaderboard for all dimensions, with respect to the inference of the anomalous diffusion exponent and the classification of the diffusion model. That is, WADNet could be the part of state-of-the-art techniques to decode the AnDi database. Moreover, the results show that WADNet can effectively extract information from noisy trajectories and vastly eliminate the undesirable influences of localization noise. The features learned by WADNet have the potential to be utilized as the quantitative representation of complex anomalous diffusion dynamics. In particular, the main architectures of WADNet for different tasks are nearly the same, where modifications are only applied on the input and output dimensions of network. Thus, our findings bring new insights into the generalization of anomalous diffusion, and could accelerate the design of a versatile strategy for single trajectory characterization.

\section*{Acknowledgements}

We thank the organizers of the AnDi Challenge for providing the database of simulated trajectories. This work is supported by the Fundamental Research Funds for the Central Universities.

\section*{References}

\bibliographystyle{unsrt}

\end{document}